\documentclass[letterpaper, 10 pt, conference]{ieeeconf}

\IEEEoverridecommandlockouts                              

\title{\LARGE \bf
Geometric Static Modeling Framework for Piecewise-Continuous Curved-Link Multi Point-of-Contact Tensegrity Robots
}

\author{Lauren Ervin and Vishesh Vikas$^{1}$
\thanks{*The material contained in this document is based upon work supported in part by a National Aeronautics and Space Administration (NASA) grant or cooperative agreement. Any opinions, findings, conclusions, or recommendations expressed in this material are those of the authors and do not necessarily reflect the views of NASA. This work was supported through a NASA grant awarded to the Alabama/NASA Space Grant Consortium.}
\thanks{$^{1}$Lauren Ervin and Vishesh Vikas are with the Agile Robotics Lab, University of Alabama, Tuscaloosa, AL 35487, USA
        {\tt\small lefaris@crimson.ua.edu, vvikas@ua.edu}}%
}

\usepackage{graphics} 
\usepackage{epsfig} 
\usepackage{mathptmx} 
\usepackage{times} 

\let\labelindent\relax
\usepackage{enumitem}
\usepackage{amsmath}  
\usepackage{amssymb}  
\usepackage{bm,soul,xcolor}
\usepackage{cite}
\usepackage{tikz}
\usetikzlibrary{arrows.meta, automata, positioning, quotes}
\usepackage{comment}
\usepackage{textcomp, gensymb}
\usepackage{array, multirow, multicol}
\usepackage{diagbox}
\usepackage{float}
\usepackage[justification=centering]{subfig}

\graphicspath{{Figures/}}
\newcommand{\Fig}{Fig. }

\newcommand{\Tab}{Tab. }
\newcommand{\Sec}{Sec. }
\newcommand{\zb}{\bm{z}_b}
\newcommand{\qoneB}{\bm{q}_1}
\newcommand{\qtwoB}{\bm{q}_2}

\renewcommand{\Re}{\mathbb{R}}
\newcommand{\Texplorer}{TeXploR}

\newcommand{\TexPlorerFullNew}{Tensegrity eXploratory Robot}

\newcommand{\smallso}{\mathfrak{so}(3)}
\newcommand{\smallse}{\mathfrak{se}(3)}
\newcommand{\bigse}{{SE}(3)}

\newcommand{\ctheta}[1]{c_{\theta_{#1}}}

\newcommand{\EulerPath}{$A1 \rightarrow C2 \rightarrow B1 \rightarrow D2 \rightarrow A1 \rightarrow B2 \rightarrow C1 \rightarrow A2 \rightarrow D1 \rightarrow B2 \rightarrow B1 \rightarrow A2 \rightarrow A1$}              

\begin{document}

\maketitle
\thispagestyle{empty}
\pagestyle{empty}

\begin{abstract}
Tensegrities synergistically combine tensile (cable) and rigid (link) elements to achieve structural integrity, making them lightweight, packable, and impact resistant. Consequently, they have high potential for locomotion in unstructured environments. 
This research presents geometric modeling of a \TexPlorerFullNew~ (\Texplorer) comprised of two semi-circular, curved links held together by 12 prestressed cables and actuated with an internal mass shifting along each link. This design allows for efficient rolling with stability (e.g., tip-over on an incline). However, the unique design poses static and dynamic modeling challenges given the discontinuous nature of the semi-circular, curved links, two changing points of contact with the surface plane, and instantaneous movement of the masses along the links.  The robot is modeled using a geometric approach where the holonomic constraints confirm the experimentally observed four-state hybrid system, proving \Texplorer~rolls along one link while pivoting about the end of the other. It also identifies the quasi-static state transition boundaries that enable a continuous change in the robot states via internal mass shifting. This is the first time in literature a non-spherical two-point contact system is kinematically and geometrically modeled. Furthermore, the static solutions are closed-form and do not require numerical exploration of the solution. The MATLAB\textregistered~simulations are experimentally validated on a tetherless prototype with mean absolute error of 4.36$\degree$.
\end{abstract}
\section{INTRODUCTION}

Robot locomotion in non-uniform terrains is a requirement for applications of space, search and rescue, and agriculture. Wheeled robots provide efficient and reliable locomotion in structured environments through continuity of contact with the surface. Their adaptation for unstructured environments include integration of legs \cite{gim_ringbot_2024} and introduction of discontinuity with compliance \cite{saranli2001rhex, quinn2003parallel}. Additionally, spherical robots have garnered interest due to the unique rolling locomotion and ability to rebound from collisions \cite{chase_review_2012, morinaga_motion_2014, ohsawa_geometric_2020}. The curved-link tensegrity robot presented in this research brings together the best of both worlds - discontinuities in wheel-designs, and locomotion ability of spherical robots. 

The term of tensegrity was introduced in the 1960s by Fuller and Snelson with initial applications in architecture \cite{snelson1996snelson}. Fundamentally, it combines rigid links under compression with prestressed cables held in \underline{ten}sion to achieve structural inte\underline{grity}. As a result, they are lightweight, packable, and do not rely on gravity to maintain structural integrity \cite{skelton2009tensegrity}. Mobile tensegrity robots aim to leverage these advantages for locomotion in unstructured environments. Commonly, these robots are based on tensegrity primitives comprised of two, three, six, and 12 rigid, straight-links \cite{paul_design_2006,sabelhaus_etal_icra_2015,caluwaerts_etal_royal_society_interface_2014}. Locomotion with straight-links is discrete and results from shape change achieved by altering cable or link lengths \cite{chen_soft_2017}. Vibration-based locomotion has also been explored by Rieffel et al \cite{rieffel2018adaptive}.

In contrast, curved link robots adapt the rolling motion of a spheres to enabling energy efficient locomotion \cite{rhodes_compact_2019, bohm_spherical_2016, bohm_dynamic_2017, kaufhold_indoor_2017}. For example, the two curved-link (2CL) robot \cite{bohm_spherical_2016} is about three times faster (normalized to body lengths/min) than the fastest straight-link counterpart. Here, the momentum change is achieved through internal mass shifting. This actuation methodology provides a greater ability of manipulating the momentum, while keeping the number of actuators low, i.e., one moving mass per link opposed to linear actuators along cables and links. The presented research adopts a similar conceptual design. Despite the design advantages, the modeling of such a two-point contact system with piecewise continuity of the curved links remains a challenge. For example, B\"{o}hm et al \cite{bohm_dynamic_2017} modeled the system where it is assumed that the robot performs uniaxial rolling about one end of the curved link and the internal mass moves along a separate straight line connecting the two ends of a link. However, this non-geometric approach makes an assumption that the robot pivots about the end of one link and rolls along the another, does not consider the masses of the curved links, and the model is qualitatively validated. Kaufhold et al \cite{kaufhold_indoor_2017} experimentally study the toppling locomotion of this robot where there is change in the point of pivot. Schorr et al \cite{schorr_kinematic_2021} also use Euler angles approach to modeling the kinematics of a three curved-link robot capable of shape morphing. In short, this approach, while effective, faces challenges with generalization and scaling, and also does not provide insight into the hybrid nature of the robot that is experimentally observed. Antali et al \cite{antali_slippingrolling_2022} propose an approach to model of two point of contact locomotion of a sphere along two orthogonal planes similar to the edge of a wall. This paper addresses the mentioned shortcomings by adopting an elegant, generalizable, and scalable geometric modeling approach that proves the hybrid state of the robot and is experimentally validated.

\textbf{Contributions.} The paper statically models a mobile, curved-link tensegrity robot using a geometric  representation and provides subsequent simulation and validation. The generalizable modeling framework is adaptable to robots with multiple points of contact and different morphologies. The model analytically proves the hybrid nature of the system where the robot exists in four states. Each state corresponds to  the motion where the robot instantaneously pivots about the end of one curved link and rolls along the other.  Static simulations provide input-output relationship between the internal mass position and ground contact points, equivalently, the robot orientation. A tetherless robot prototype experimentally validates the static model with high accuracy.

\textbf{Paper Organization} The next section explores the kinematics of the robot and proves the hybrid nature of the system. The static modeling is analyzed in the following section. Thereafter, the fabrication and design methodology of the tensegrity robot is presented. Next, simulations with a scenario of state change along the transition boundaries for a quasi-static control path and real-world experimental results are discussed. The final section concludes the research and explores the future directions.
\section{Robot Kinematics}


\begin{figure}[!ht]
    \centering
    \includegraphics[width = 0.8\columnwidth]{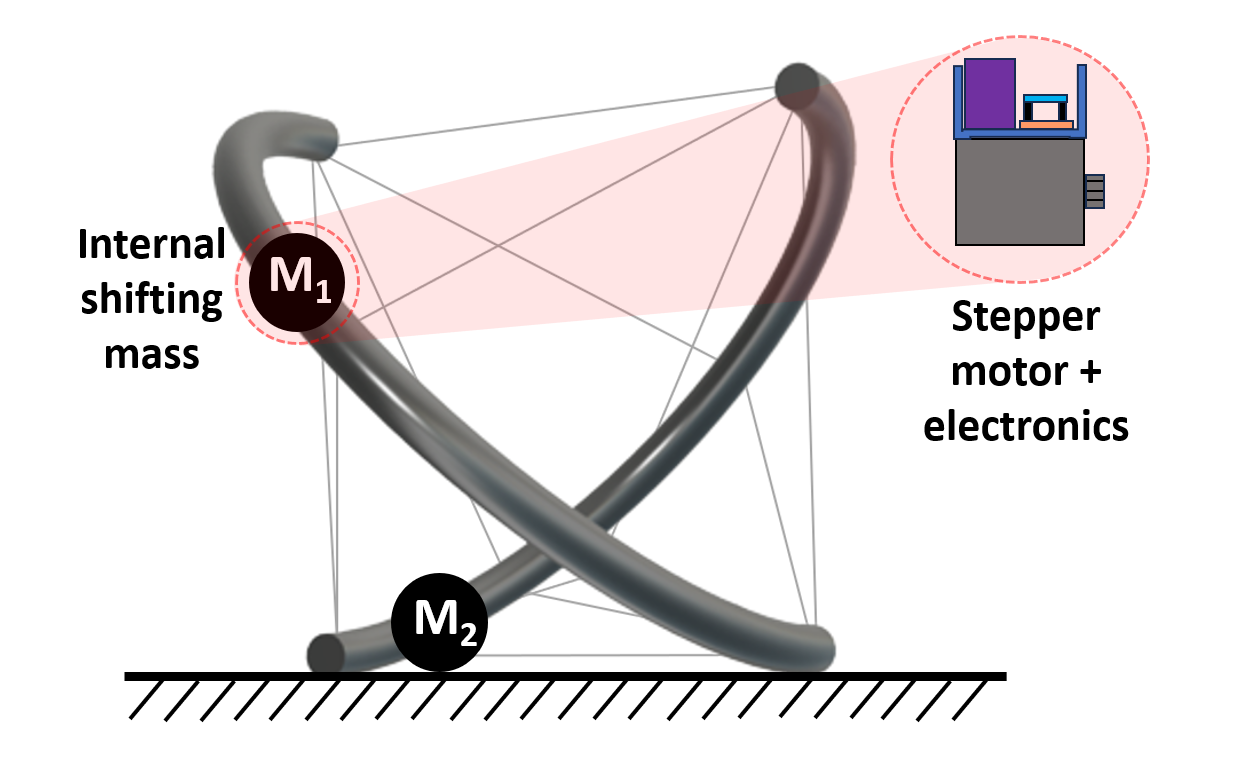}
    \caption{Two curved-link tensegrity robot, \Texplorer,~is structurally held together with 12 tensile cables. Change of robot pose is achieved through internal masses shifting along the curved links. At any instant in time, it has two points of contact with the ground.}
    \label{fig:texplor}
\end{figure}

\subsection{Problem definition and notation}
\begin{figure}[!ht]
    \centering
    \includegraphics[width = \columnwidth]{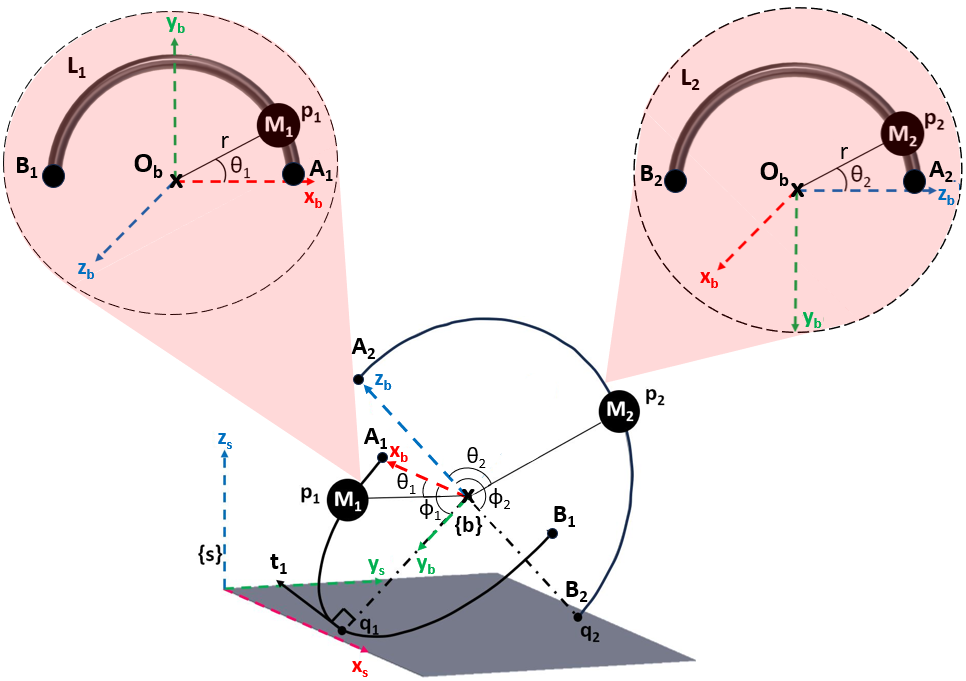}
    \caption{Geometric relationship the spatial $\{s\}$, body $\{b\}$ and link $\{1\},\{2\}$ coordinate systems. Here, $\{b\},\{1\}$ coincide and $\bm{z}_s$ is the normal to the surface of locomotion. The internal masses $P_i$ and ground contact points $Q_i$ along the links are denoted using $\theta_i,\phi_i$ respectively where $i=1,2$. The tangents $\bm{t}_i$ at the contact points lie along the surface.}
    \label{fig:axes}
\end{figure}

The \TexPlorerFullNew~(\Texplorer), \Fig \ref{fig:texplor}, comprises of two curved-links held together with a tensile cable with two motor assemblies that are free to move along the links. The two links, $\{L_1,L_2\}$, are modeled as semi-circular arcs with radii $r$ with masses $\{m_1,m_2\}$, \Fig \ref{fig:axes}. The two endpoints of the two arcs are denoted as $\{A_i,B_i\}$ for link $L_i$. For the remainder of this section, $i=1,2$ corresponding to the two links. The body coordinate system $\{b\}$ is fixed on the robot body with origin $O_b$ and orthonormal basis $U_b =\left[\bm{x}_b,\bm{y}_b,\bm{z}_b\right]\in SO(3)$. Let coordinate systems $\{1,2\}$ be fixed on the corresponding link with origin at the center of each arc, such that the $x$ and $z$ axes respectively are along the arc end points $B_iA_i$, and normal to the arc plane. The center of the two links are coincident and $\{1\}$ coincides with $\{b\}$. The inertial coordinate system $\{s\}$ is fixed at origin $O_s$ with orthonormal basis $U_s=\left[\bm{x}_s,\bm{y}_s,\bm{z}_s\right]\in SO(3)$ such that $\bm{z}_s$ is normal to the plane of locomotion.
The motor assemblies are modeled as point masses $M_{i}$ that move along the corresponding link. These point masses $P_i$ are defined using angle $\theta_i$ from $\bm{x}_i$ and denoted by $\bm{p}_i\in \Re^{3\times 1}$. The instantaneous points of contact, $Q_i$, between the links and the ground plane are represented by $\bm{q}_i\in\Re^{3\times 1}$ and parameterized using ground contact angles $\phi_i$ from $\bm{x}_i$. The tangents of the arc at these instantaneous points of contact are denoted by $\bm{t}_i\in\Re^{3\times 1}$. Terminology of the shifting masses and points of contact is summarized in \Tab \ref{tab:arcs}.
\begin{table}[h]
\begin{center}
\caption{Shifting Masses and Points of Contact}
\label{tab:arcs}
\begin{tabular}{|c|c|c|c|}
\hline
\multicolumn{2}{|c|}{Motor $i=1,2$} & %
\multicolumn{2}{c|}{Point of contact $i=1,2$}\\
\hline
$\bm{p}_i$     &  Position of motor $i$ &
$\bm{q}_i$     &  Link $i$ \\ 
$\theta_i$     &  Angle along motor &
$\phi_i$     &  Angle link $i$\\
$M_i$ & Shifting mass &
$\bm{t}_i$ & Tangent line $i$ \\\hline \hline
$r$& Arc radius & $R_{12}$ & Rotation matrix b/w $
\{1\}, \{2\}$\\\hline
$A_1, B_1$ & Arc 1 endpoints & $A_2, B_2$ & Arc 2 endpoints\\\hline
\end{tabular}
\end{center}
\vspace{-10pt}
\end{table}

Ultimately, it is desired to find the transformation matrix $T_{sb}\in SE(3)$, such that
\begin{align}
    T_{sb} = \begin{bmatrix}
        R_{sb} & \bm{o}_{sb}\\ \bm{0}_{3\times 1}& 1
    \end{bmatrix}
\end{align}
where $R_{sb}\in SO(3)$, $\bm{o}_{sb}$ are the rotation matrix and displacement between origins of $\{s\},\{b\}$.

\subsection{Kinematics}
The position vectors  $\bm{p}_i,\bm{q}_i$ are represented as a combination of the arc radius and respective angles along arcs.
\begin{equation}
    \bm{p}_i^i = r\begin{bmatrix}
        c_{\theta_i}\\s_{\theta_i}\\0
    \end{bmatrix}, \quad 
    \bm{q}_i^i = r\begin{bmatrix}
        c_{\phi_i}\\s_{\phi_i}\\0
    \end{bmatrix}    
\end{equation}
where $c_{\alpha_i}=\cos(\alpha_i), s_{\alpha_i}=\sin(\alpha_i), \forall i=1,2, \alpha =\theta,\phi$, and the superscript $i$ refers to the c.s. of representation. The transformation matrix $T_{12}\in \bigse$ changes the representation between c.s. $\{2\}$ to $\{1\}$. That is, for some vector $\bm{v}\in\Re^{3\times 1}$ 
\begin{align}
\begin{gathered}
\begin{bmatrix}\bm{v}^1\\1\end{bmatrix} = T_{12}\begin{bmatrix}\bm{v}^2\\1\end{bmatrix} \Rightarrow \bm{v}^1 = \bm{o}_{12}^1 + R_{12}\bm{v}^2\\
T_{12}  = \begin{bmatrix}
    R_{12} & \bm{o}_{12}\\
    \bm{0}_{1\times 3} & 1
\end{bmatrix}, R_{12} = \begin{bmatrix}
0 & 0 & 1 \\
0 & -1 & 0 \\
1 & 0 & 0 \\
\end{bmatrix}, \bm{o}_{12}^1 =  \begin{bmatrix} 0\\0\\0\end{bmatrix} 
\end{gathered}
\label{Eqn:T12def}
\end{align}
where $R_{12}\in SO(3)$ and $\bm{o}_{12}\in \Re^{3\times 1}$ are the rotation matrix and the displacement between the origins of the $\{1\},\{2\}$. For the remainder of the paper, we disregard the superscript for the vector notation, as all the representations will be in the robot body c.s. $\{b\}$ which coincides with $\{1\}$. Hence,
\begin{equation}
\begin{gathered}
\bm{p}_1 = r\begin{bmatrix}
        \ctheta{1}\\s_{\theta_1}\\0
    \end{bmatrix}, \quad
\bm{p}_2= \bm{o}_{12} + R_{12}\bm{p}_2^2 = 
        r \begin{bmatrix} 0\\
-s_{\theta_2}\\
c_{\theta_2} \end{bmatrix} \\
\bm{q}_1 = r\begin{bmatrix}
        c_{\phi_1}\\s_{\phi_1}\\0
    \end{bmatrix}, \quad 
\bm{q}_2= \bm{o}_{12} + R_{12}\bm{q}_2^2 = r\begin{bmatrix} 0\\
-s_{\phi_2}\\
c_{\phi_2} \end{bmatrix}
\end{gathered}
\end{equation}
The `free-vector' tangent $\bm{t}_i$ at points of contact $Q_i$ are
\begin{align}
    \bm{t}_i = \frac{\partial \bm{q}_i}{\partial \phi_i}
\end{align}

\subsection{Rolling constraint}

Pure rolling without slipping is assumed at the points of contact, $Q_1,Q_2$. Hence, they have zero velocity, i.e., $\bm{\dot{q}}_1=\bm{\dot{q}}_2=0$. {We use the Lie group approach for modeling and follow the notation as per}\cite{murray_mathematical_2017}. The body twist, $\xi_b \in \mathbb{R}^6$, encodes the angular velocity $\omega_b$ and linear velocity of the origin $O_b$, $v_b$ expressed in $\{b\}$. The hat operator $~\hat{}~$ maps $
\Re^{6\times 1} \rightarrow \smallse$ and $\Re^{3\times 1} \rightarrow \smallso$
\begin{equation}
\xi_b = \begin{bmatrix}
    \omega_b\\v_b
\end{bmatrix}, \quad    \widehat{\xi}_b = \begin{bmatrix}\widehat{\omega}_b & v_b\\
0 & 0 \end{bmatrix} \in \smallse \nonumber
\end{equation}
Following this notation, the rolling constraint is
\begin{equation}
    \bm{\dot{q}}_i = v_b + \hat{\omega}_b\bm{q}_i = 0 \nonumber
\end{equation}
In this paper, we restrict ourselves to static analysis and do not examine this constraint given the modeling complexity, especially, the hybrid nature of such two-point contact system to be discussed next.
\subsection{Holonomic constraint}\label{holonomic}
It is assumed that the robot is in physical contact with the surface of locomotion at all times. This implies that the two points $Q_1,Q_2$ are in contact with the ground and the tangents at those points also lie on the ground plane. The ground plane is defined using the surface normal which is the z-axis $\bm{z}_s$ of $\{s\}$.  Another kinematic interpretation of this is that the $\bm{z_s}$ component of position $\bm{q_i}$ is the minimum amongst all the points on the link. For the two points of contact $Q_{i}$
\begin{align}
    \begin{gathered}
    \relax[\bm{z_s}^T,1] \cdot \left( T_{sb}\begin{bmatrix}
        \bm{q}_i\\1    \end{bmatrix}\right)=0,\quad 
    [\bm{z_s}^T,1] \cdot \left( T_{sb}\begin{bmatrix}
        \bm{t}_i\\0    \end{bmatrix}\right)=0 
    \end{gathered}
\end{align}
This can be simplified to 
\begin{align}
    \begin{gathered}
    \begin{bmatrix}
        (\bm{q}_1-\bm{q}_2)^T\\        \left(\bm{t}_1\right)^T\\        \left(\bm{t}_2\right)^T
    \end{bmatrix}\bm{z_b} = \bm{0}, \quad
        \bm{z_b} = R_{sb}^T\bm{z_s} 
    ,~ \mathrm{s.t.}\quad 
    \bm{z}_s = \begin{bmatrix}
        0\\0\\1
    \end{bmatrix}
        \label{eq:6}
    \end{gathered}
\end{align}

Geometrically, this implies that vectors $(\qoneB-\qtwoB), \bm{t}_1,\bm{t}_2$ lie in the same plane and the vector $\bm{z_b}$ is the surface normal. However, there is discontinuity at the edges of the links, i.e., $\phi_i=0,180\degree$ where the tangents cannot be defined. These discontinuities can be separated into three unique cases:\vspace{5pt}

\textbf{Case 1: Tangent vector $\bm{t}_2$ not defined.} Here, vector $\bm{z}_b$ is normal to the  vector $Q_1Q_2$ and tangent vector at $Q_1$. The direction of $\bm{z}_b$ can be deduced from the fact that $\bm{o}_{sb}$, origin $O_b$ of $\{b\}$ lies above the ground plane, i.e., $\bm{z}_s^T\bm{o}_{sb}>0$, and $Q_1,Q_2$ are in contact with the ground. Mathematically,
\begin{align*}
    \bm{z}_s^T\left(\bm{o}_{sb} +R_{sb}\bm{q}_i\right) = 0,~\mathrm{and}~\bm{z}_s^T\bm{o}_{sb}>0 ~ \Rightarrow ~-(\underbrace{R_{sb}^T\bm{z}_s}_{\bm{z}_b^T})^T\bm{q}_i>0
\end{align*}
The analytical solution for $\zb$ can then be summarized as
\begin{align}
\begin{gathered}
\bm{\zb}=\pm \frac{\bm{t}_1\times(\bm{q}_1-\bm{q}_2)}{|\bm{t}_1\times(\bm{q}_1-\bm{q}_2)|} \quad \mathrm{s.t.}\quad -\zb^T\qoneB=-\zb^T\qtwoB>0\\
\bm{\zb} = -\frac{1}{\sqrt{2}}[c_{\phi_1},s_{\phi_1},c_{\phi_2}]^T, \quad
    -\zb^T\qoneB=-\zb^T\qtwoB=\frac{r}{\sqrt{2}}
    \end{gathered} \label{Eqn:holonomicCase1}
 \end{align}
\textbf{Case 2: Tangent vector $\bm{t}_1$ not defined.} 
Similar to the previous case, $\bm{z}_b$ is normal to the vector $Q_1Q_2$ and the tangent vector at $Q_2$. Using the surface-contact constraint, the analytical expression of unique $\bm{z}_b$ is
\begin{align}
\begin{gathered}
\bm{\zb}=\pm\frac{\bm{t}_2\times(\bm{q}_2-\bm{q}_1)}{|\bm{t}_2\times(\bm{q}_2-\bm{q}_1)|} \quad \mathrm{s.t.}\quad -\zb^T\qoneB=-\zb^T\qtwoB>0\\
\bm{\zb} = \frac{1}{\sqrt{2}}[-c_{\phi_1},s_{\phi_2},-c_{\phi_2}]^T, \quad
    -\zb^T\qoneB=-\zb^T\qtwoB=\frac{r}{\sqrt{2}}
\end{gathered}
\label{Eqn:holonomicCase2}
\end{align}

\noindent \textbf{Case 3: Both tangent vectors $\bm{t}_i$ defined.} For this case, $\bm{z}_b$ is normal to the plane containing vector $Q_1Q_2$ and tangent vectors at both points $Q_1$ and $Q_2$. Consequently, the angles $\phi_i$ can be analytically written as
\begin{align}
\begin{gathered}
    (\bm{q}_1-\bm{q}_2)^T(\bm{t}_1\times\bm{t}_2) = 0 \\
    \Rightarrow -r^3(s_{\phi_1}+s_{\phi_2}) = 0 \Rightarrow {\phi_2 = -\phi_1}\mathrm{~or~}180\degree+\phi_2\\
    \bm{\zb}=\frac{\bm{t}_1\times\bm{t}_2}{|\bm{t}_1\times\bm{t}_2|} \\
    \phi_2 = \left\{\begin{array}{c@{,}c}
    -\phi_1 & \quad \bm{\zb} = \frac{1}{\sqrt{2c_{\phi_1}^2 +s_{\phi_1}^2}}[c_{\phi_1},s_{\phi_1},c_{\phi_1}]^T\\
    180\degree+\phi_1& \quad \bm{\zb} = \frac{1}{\sqrt{2c_{\phi_1}^2 +s_{\phi_1}^2}}[c_{\phi_1},s_{\phi_1},-c_{\phi_1}]^T
    \end{array}\right.
    \end{gathered}
\end{align}
This case is infeasible due to the range constraints on $\phi_1,\phi_2$. More specifically, for $\phi_2 > 0$, both $-\phi_1$ and $180\degree + \phi_1$ are outside of the solution space, i.e., the arc range $\phi_2\in[0,180\degree]$. 

\subsection{Hybrid system} \label{hybrid}
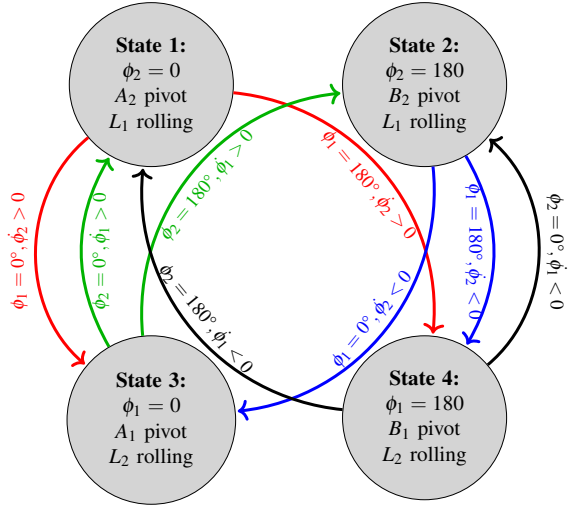
\begin{figure}[!ht]
\centering
\scalebox{0.8}{\begin{tikzpicture}[shorten >=1pt,node distance=4cm, auto]
  \tikzstyle{every state}=[fill={rgb:black,2;white,10}]
  \node[state, text width=1.8cm, align=center] (s_1)  {\textbf{State 1:} $\phi_2 = 0$ $A_2$ pivot  $L_1$ rolling};
  \node[state]           (s_2) [right=1.8cm of s_1, text width=1.8cm, align=center]     {\textbf{State 2:} $\phi_2 = 180$  $B_2$ pivot $L_1$ rolling};
  \node[state] (s_3) [below=2.8cm of s_1, text width=1.9cm, align=center] {\textbf{State 3:} $\phi_1 = 0$  $A_1$ pivot $L_2$ rolling};
  \node[state]           (s_4) [below=2.8cm of s_2, text width=1.8cm, align=center]     {\textbf{State 4:} $\phi_1 = 180 $  $B_1$ pivot $L_2$ rolling};

  \path[->]
  (s_1) edge [red, bend left=45, line width=0.5mm]  node[sloped, below, node font=\small] {$\phi_1=180\degree, \dot{\phi_2} > 0$}  (s_4)
        edge [red, bend right=50, line width=0.5mm]  node[sloped, above, node font=\small] {$\phi_1=0\degree, \dot{\phi_2} > 0$} (s_3)
  (s_2) edge [blue, bend left=45, line width=0.5mm]  node[sloped, above, node font=\small] {$\phi_1=0\degree, \dot{\phi_2} < 0$} (s_3)
        edge [blue, bend left, line width=0.5mm]  node[sloped, below, node font=\small] {$\phi_1=180\degree, \dot{\phi_2} < 0$} (s_4)
  (s_3) edge [green!70!black, bend left=45, line width=0.5mm] node[sloped, below, node font=\small] {$\phi_2=180\degree, \dot{\phi_1} > 0$} (s_2)
        edge [green!70!black, bend left, line width=0.5mm]  node[sloped, below, node font=\small] {$\phi_2=0\degree, \dot{\phi_1} > 0$} (s_1)
  (s_4) edge [bend left=45, line width=0.5mm] node[sloped, above, node font=\small] {$\phi_2=180\degree, \dot{\phi_1} < 0$} (s_1)
        edge [bend right=50, line width=0.5mm]  node[sloped, above, node font=\small, rotate=180] {$\phi_2=0\degree, \dot{\phi_1} < 0$} (s_2);
\end{tikzpicture}}
\caption{Four state hybrid system model of the \Texplorer. The state transition is decided by the pivot and the rate of change of point of contact $\dot{\phi}_i$. During each state, the robot pivots about one of the ends of the curved links while rolling about the other.}
\label{fig:FSM}
\end{figure}

Given that the surface normal is defined only for cases 1 and 2, the robot can be modeled as a hybrid system transitioning between four states. Each case corresponds to the robot instantaneously pivoting about about one of the end points of the curved link.
\begin{enumerate}[wide, labelwidth=0pt, labelindent=0pt]
    \item[] \textbf{State 1}: $\phi_1\in (0,180\degree), \phi_2=\phantom{1}0\degree,\phantom{8}$ $A_2$ pivot, roll along $L_1$
    \item[] \textbf{State 2}:  $\phi_1\in (0,180\degree), \phi_2=180\degree$, $B_2$ pivot, roll along $L_1$
    \item[] \textbf{State 3}: $\phi_1=\phantom{1}0\degree\phantom{8}, \phi_2\in (0,180\degree)$, $A_1$ pivot, roll along $L_2$ 
    \item[] \textbf{State 4}: $\phi_1=180\degree, \phi_2\in (0,180\degree)$, $B_1$ pivot, roll along $L_2$ 
\end{enumerate}
As an example, the geometric representation of the robot in State 2 and pivoting about the link end point $B_2$ where tangent vector $\bm{t}_2$ is not defined is shown in \Fig \ref{fig:axes}. The robot is a hybrid system transitions between four states of locomotion. More importantly, all the states are not connected to each other. For example, the robot cannot transition directly from State 1 to 2, instead the transition is $1\rightarrow 3 \rightarrow 2$ or $1\rightarrow 4 \rightarrow 2$. In each of these states, the transition occurs when $\phi_i$ is fixed at $0\degree, 180\degree$ and $\dot{\phi}_{i\pm1} >0$. The hybrid state system is summarized in \Fig \ref{fig:FSM}. 

Let's analyze the transition between States $1 \rightarrow 3 \rightarrow 2\rightarrow 4$. While in State 1, $\phi_1$ reaches the end of $L_1$, point $A_1$. The robot transitions to State 3 as $\phi_1\rightarrow 0\degree$ and pivots about $A_1$ while $\phi_2$ begins to traverse the entirety of $L_2$. Next, it arrives at the $\phi_2 = 180\degree$, i.e., $B_2$. Now, with $\dot{\phi_1} > 0$, and the robot transitions to State 2. Here, $\phi_2 = 180\degree$ as $\phi_1 \in (0\degree, 180\degree$) traverses $L_1$, stopping at the $B_1$. Finally, it transitions to State 4. The robot repeatedly transitions between these four states either forwards or backwards for a rolling sequence.

\subsection{Generalizability for different robot morphologies}
The extension of this framework to different curved-link tensegrity morphologies will require it to be applied to systems with variations in robot shape, i.e., $T_{12}$, number of curved links, length of the curved links.

\begin{figure}[!ht]
    \centering
    \includegraphics[width = \columnwidth]{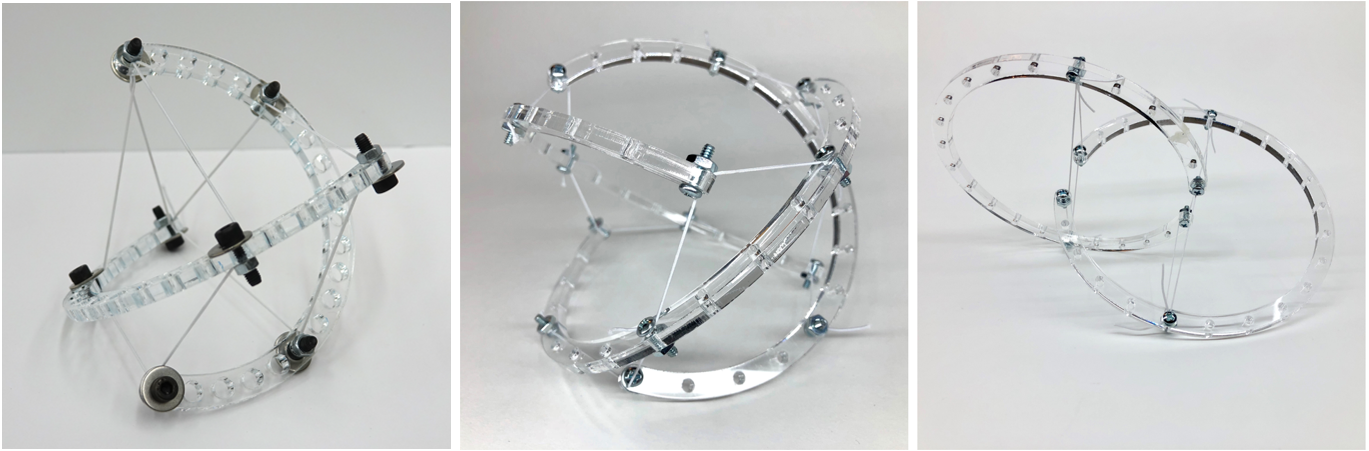}
    \subfloat(a){\hspace{.33\linewidth}}
    \subfloat(b){\hspace{.33\linewidth}}
    \subfloat(c){\hspace{.33\linewidth}}
    \caption{Different robot morphologies that can be modeled with the generalizable framework: (a) A two link prototype with a different shape and $T_{12}$. (b) A three curved-links resulting in three points of ground contact. (c) A two link prototype with arc length more than $180\degree$.}
    \label{fig:generalizable}
\end{figure}

\textbf{Shape morphing.} The variations in cable tensions have direct impact on the robot shape, more specifically, the transformation matrix $T_{12}$ as defined in \eqref{Eqn:T12def}. It has been documented that variations in the orientation and distance between the two curved links has direct impact on the rolling locomotion path \cite{kaufhold_indoor_2017}. For this research, the curved links are orthogonal with coincident arc centers. This enables a straight trajectory of locomotion during a rolling sequence. Shape morphing of the robot, reflected in $T_{12}$ and shown in \Fig \ref{fig:generalizable}(a), will allow the robot to turn. This change can be incorporated by updating the point of contact $Q_2$ and the tangent vector as a function of $\bm{o}_{12},R_{12}$. 
\begin{align*}
    \bm{q}_2 = \bm{o}_{12} + R_{12}\bm{q}_2^2, \quad \bm{t}_2 = \frac{\partial \bm{q}_2}{\partial \phi_2}
\end{align*}

\textbf{Number of curved links.} The framework is adaptable to tensegrity robots with more curved links, e.g., three as shown in \Fig \ref{fig:generalizable}(b). For this morphology, \eqref{eq:6} will be modified to 
\begin{align}
    \begin{gathered}
    \begin{bmatrix}
        (\bm{q}_1-\bm{q}_2) & 
        (\bm{q}_2-\bm{q}_3) &
        \left(\bm{t}_1\right)&        \left(\bm{t}_2\right)&
        \left(\bm{t}_3\right)
    \end{bmatrix}^T\bm{z_b} = \bm{0}
    \end{gathered}
\end{align}
where $\bm{q}_3, \bm{t}_3$ are defined for the new point of contact $Q_3$ on the third curved link. Consequently, the analysis will result in a change in number of robot states in the hybrid system. 

\textbf{Curve link length.} The arc lengths of the curves can be varied beyond $180\degree$ as illustrated in \Fig \ref{fig:generalizable}(c). This alteration in the robot morphology will result in reconsideration of the possibility of a solution existing for `Case 3' discussed in \Sec \ref{holonomic}. This may result in an increase in the number of robot states from the four discussed earlier.
\section{Static Modeling}\label{staticmodeling}
The two-point of contact and hybrid nature of the system results in a robot existing in four states. Consequently, the analytical solution to the statics problem is obtained by equating the wrench to zero and solving for the unique point of ground contact corresponding to the robot orientation.

\subsection{Closed form solution for point of contact}
\begin{figure}[!ht]
    \centering
    \includegraphics[width = 0.8\columnwidth]{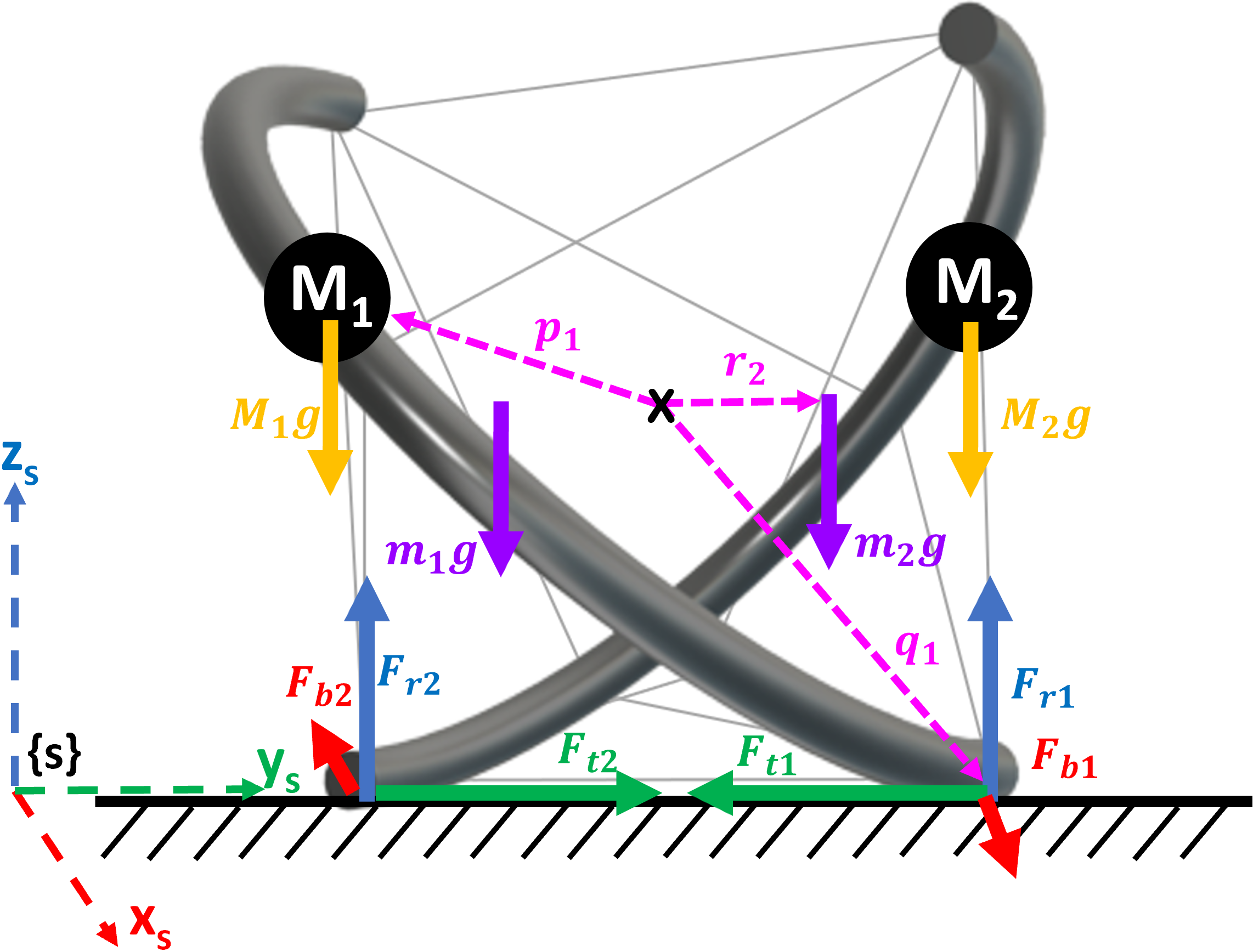}
    \caption{The free-body diagram of \Texplorer with the intertial and reaction forces.}
    \label{fig:fbd}
    \vspace{-25pt}
\end{figure}

The wrench, $\mathcal{F}$, is comprised of the total force, $f$, and moment, $m$, acting on the body. For the static formulation, the wrench $\mathcal{F}$ about $O_b$ is equated to zero. The corresponding free body diagram of the two point of contact system showing all forces is shown in \Fig \ref{fig:fbd}. Here, the center of mass of the curved links $\bm{r}_1=\bm{r}_2 = [0,{2r}/{\pi} , 0]^T$.

\begin{align*}
\begin{gathered}   
    f = \begin{bmatrix}
    F_{b1}-F_{b2}\\
    F_{t1}-F_{t2}\\
        (M_1+M_2+m_1+m_2)g - (F_{r1} +F_{r2})
    \end{bmatrix}\\
    m =\begin{array}{l}\left(M_1g\hat{\bm{p}}_1 + M_2g\hat{\bm{p}}_2 + m_1g\hat{\bm{r}}_1 + m_2g\hat{\bm{r}}_2 - F_{r1}\hat{\bm{q}}_1 - F_{r2}\hat{\bm{q}}_2\right)\bm{\zb}\\
    +\left(F_{b1}\hat{\bm{q}}_1 - F_{b2}\hat{\bm{q}}_2\right)\bm{x}_b + \left(F_{t1}\hat{\bm{q}}_1 - F_{t2}\hat{\bm{q}}_2\right)\bm{y}_b
    \end{array}
\end{gathered}
\end{align*}
These are six equations with seven unknowns $F_{bi},F_{ti},F_{ri}$ and $\phi_i$ along the rolling link for the corresponding state. These can be simplified to four equations with three unknowns $F_{r1},F_{r2},\phi_i$ by observing that $F_{b1}=F_{b2}, F_{t1}=F_{t2}$, and $Q_1Q_2$ lies in the same plane as $\bm{x}_s,\bm{y_s}$, i.e., $(\hat{\bm{q}}_1-\hat{\bm{q}}_2)\bm{x}_b=(\hat{\bm{q}}_1-\hat{\bm{q}}_2)\bm{y}_b=0$. Hence, 
\begin{align*}
\begin{gathered}
        \begin{bmatrix}
        (M_1+M_2+m_1+m_2)g - (F_{r1} +F_{r2})\\
        \left(M_1g\hat{\bm{p}_1} + M_2g\hat{\bm{p}_2} + m_1g\hat{\bm{r}_1} + m_2g\hat{\bm{r}_2} - F_{r1}\hat{\bm{q}_1} - F_{r2}\hat{\bm{q}_2}\right)\bm{\zb}
    \end{bmatrix} = 0
\end{gathered}
\end{align*}

Using the calculated $\zb$ from \Sec \ref{holonomic} for each state, the ground contact angles and the reaction forces an be analytically determined as

\noindent\textbf{State 1:} Pivot about $A_2$ ($\phi_2 = 0\degree$), rolling on $L_1$
    \begin{align} \label{Eqn:State1}
    \begin{gathered}
        \bm{\zb} = -\frac{1}{\sqrt{2}}[c_{\phi_1},s_{\phi_1},1]^T, \quad \tan\phi_1 = \frac{s_{\theta_1}-s_{\theta_2}}{c_{\theta_1}}\\
        F_{r1} = Mg\left[-1 + \frac{c_{\phi_1 + \theta_2}}{2} - c_{\theta_2} + c_{\phi_1 - \theta_1} - \frac{c_{\phi_1 - \theta_2}}{2}\right] - mg\\
        F_{r2} = Mg\left[-1 - \frac{c_{\phi_1 + \theta_2}}{2} + c_{\theta_2} - c_{\phi_1 - \theta_1} + \frac{c_{\phi_1 - \theta_2}}{2}\right] - mg\\
    \end{gathered}
    \end{align}

\noindent\textbf{State 2:} Pivot about $B_2$ ($\phi_2 = 180\degree$), rolling on $L_1$
    \begin{align} \label{Eqn:State2}
    \begin{gathered}
        \bm{\zb} = -\frac{1}{\sqrt{2}}[c_{\phi_1},s_{\phi_1},1]^T, \quad \tan\phi_1 = \frac{s_{\theta_1}-s_{\theta_2}}{c_{\theta_1}}\\
        F_{r1} = Mg\left[-1 + \frac{c_{\phi_1 + \theta_2}}{2} + c_{\theta_2} + c_{\phi_1 - \theta_1} - \frac{c_{\phi_1 - \theta_2}}{2}\right] - mg\\
        F_{r2} = Mg\left[-1 - \frac{c_{\phi_1 + \theta_2}}{2} - c_{\theta_2} - c_{\phi_1 - \theta_1} + \frac{c_{\phi_1 - \theta_2}}{2}\right] - mg
        %
    \end{gathered}
    \end{align}
    
\noindent\textbf{State 3:} Pivot on $A_1$ ($\phi_1 = 0\degree$), rolling about $L_2$:
    \begin{align}\label{Eqn:State3}
    \begin{gathered}
        \bm{\zb} = \frac{1}{\sqrt{2}}[-1,s_{\phi_2},-c_{\phi_2}]^T,\quad \tan\phi_2 = \frac{s_{\theta_2}-s_{\theta_1}}{c_{\theta_2}}\\
        F_{r1} = Mg\left[-1 - \frac{c_{\phi_2 + \theta_1}}{2} + c_{\theta_1} - c_{\phi_2 - \theta_2} + \frac{c_{\phi_2 - \theta_1}}{2}\right] - mg\\
        F_{r2} = Mg\left[-1 + \frac{c_{\phi_2 + \theta_1}}{2} - c_{\theta_1} + c_{\phi_2 - \theta_2} - \frac{c_{\phi_2 - \theta_1}}{2}\right] - mg
        %
    \end{gathered}
    \end{align}

\noindent\textbf{State 4:} Pivot on $A_2$ ($\phi_1 = 180\degree$), rolling about $L_2$:
    \begin{align} \label{Eqn:State4}
    \begin{gathered}
        \bm{\zb} = \frac{1}{\sqrt{2}}[-1,-s_{\phi_2},c_{\phi_2}]^T, \quad \tan\phi_2 = \frac{s_{\theta_2}-s_{\theta_1}}{c_{\theta_2}}\\
        F_{r1} = Mg\left[-1 - \frac{c_{\phi_2 + \theta_1}}{2} + c_{\theta_1} - c_{\phi_2 - \theta_2} + \frac{c_{\phi_2 - \theta_1}}{2}\right] - mg\\
        F_{r2} = Mg\left[-1 + \frac{c_{\phi_2 + \theta_1}}{2} - c_{\theta_1} + c_{\phi_2 - \theta_2} - \frac{c_{\phi_2 - \theta_1}}{2}\right] - mg\\
        %
    \end{gathered}
    \end{align}
    

    
\subsection{Rotation matrix}
As evident, the static analysis allows for closed form solution for $\bm{z}_b$. The equilibrium orientation does not provide any information about the planar position of the robot on the surface, i.e., $x,y$ components of $O_b$ are `free' while the $z$ component is fixed at $r/\sqrt{2}$ as calculated in \eqref{Eqn:holonomicCase1},\eqref{Eqn:holonomicCase2}. Hence, ignoring additional rotation about $\bm{z}_s$ the rotation matrix can be calculated using the Rodrigues' formula and vectors $\bm{z_s},\bm{z_b}$
\begin{align}
\begin{gathered}
    R_{sb}(\bm{u},\theta) = I + \sin{\theta}\hat{\bm{u}} + (1 - \cos{\theta})\hat{\bm{u}}^2\\
    \mathrm{where~}     \bm{u} = \frac{\bm{\zb}\times\bm{z}_s}{|\bm{\zb}\times\bm{z}_s|}, \quad \theta = \mathrm{atan2}\left(|\bm{\zb}\times\bm{z}_s|, \bm{\zb}^T\bm{z}_s\right)
\end{gathered}
\end{align}
Rodrigues' formula is used to convert the axis $\bm{u}$ and rotation angle $\theta$ into a rotation matrix in $SO(3)$. Additionally, $\bm{o}_{sb}^s = x,y,\sqrt{r}/2]^T$ for some $x,y$, i.e., the robot can sit anywhere on the plane. A closed form solution for $T_{sb}$ can be obtained by using the calculated $R_{sb}$ and $\bm{o}_{sb}$. 



\section{TEXPLORER DESIGN AND FABRICATION}\label{design}

\subsection{Static form finding}
The robot is a tensegrity mechanism that combines elements of tension and compression. The shape of the robot, $T_{12}$, depends upon the cable segment lengths. Form-finding determines the lengths that dictate a desired shape. These can be found by  minimizing the energy of the system \cite{tibert_review_2003, 
connelly_second-order_1996}. We define the problem in terms of the generalized coordinates, i.e., the screw $\xi$ using a Lie groups approach. The four connection points on each curved link are identical and expressed in the corresponding link coordinate systems. Consequently, the 12 cable segments are expressed as $\bm{d}\in\Re^{12\times 4}$ and the $i$th cable $\bm{d}_i$ is denoted by the $i$th row
\begin{align}
\begin{gathered}
    \bm{d}(\xi)  = PC_1- e^{\widehat{\xi}}PC_2= PC_1- T_{12}PC_2\\
    P = \begin{bmatrix}
        \bm{p}_{A} & \bm{p}_{B} & \bm{p}_{C} & \bm{p}_{D} \\ 1 & 1 & 1 & 1
    \end{bmatrix} = \begin{bmatrix}
        -r & 0 & 0 & 1\\
        \frac{-r}{2} & \frac{-\sqrt{3}r}{2} & 0 & 1\\
        \frac{r}{2} & \frac{\sqrt{3}r}{2} & 0 & 1\\
        r & 0 & 0 & 1
    \end{bmatrix}^T
\end{gathered}
\end{align}
where $P$ are the concatenated homogeneous representation of points $A,B,C,D$ on each arc, and the \Fig \ref{fig:cab}. The connectivity matrices $C_1,C_2$ mathematically represent the two end-points each cable
\begin{align*}
    C_i[j,k] = \left\{ \begin{array}{c@{,~}c}
     1 &  \mathrm{if~cable~k~contains~vertex~j~\mathrm{on~link}~i}\\
     0 & \mathrm{otherwise}
    \end{array}\right.
\end{align*}
Now, the energy minimization problem can be written as
\begin{align}
    \xi^* = \min_{\xi} \sum_{i=1}^{12}\frac{1}{2}k_i\left(|\bm{d}_i|-d_{0,i}\right)^2
\end{align}
where $k_i,d_{0,i}$ are the stiffness and the free length of the corresponding cable. There are two different free length cable segments in the system corresponding to edge-to-edge and edge-to-middle segments, $d_{0,1} = 3.25"$ and $d_{0,2} = 3"$. The optimal $T_{12}=e^{\widehat{\xi}}$ corresponding to the free-length matches \eqref{Eqn:T12def} where $d_1=5.8551, d_2=5.5412$, visualized in \Fig \ref{fig:cab}(a).

\subsection{Tendon routing}
\begin{figure}[!ht]
    \centering
    \includegraphics[width = \linewidth]{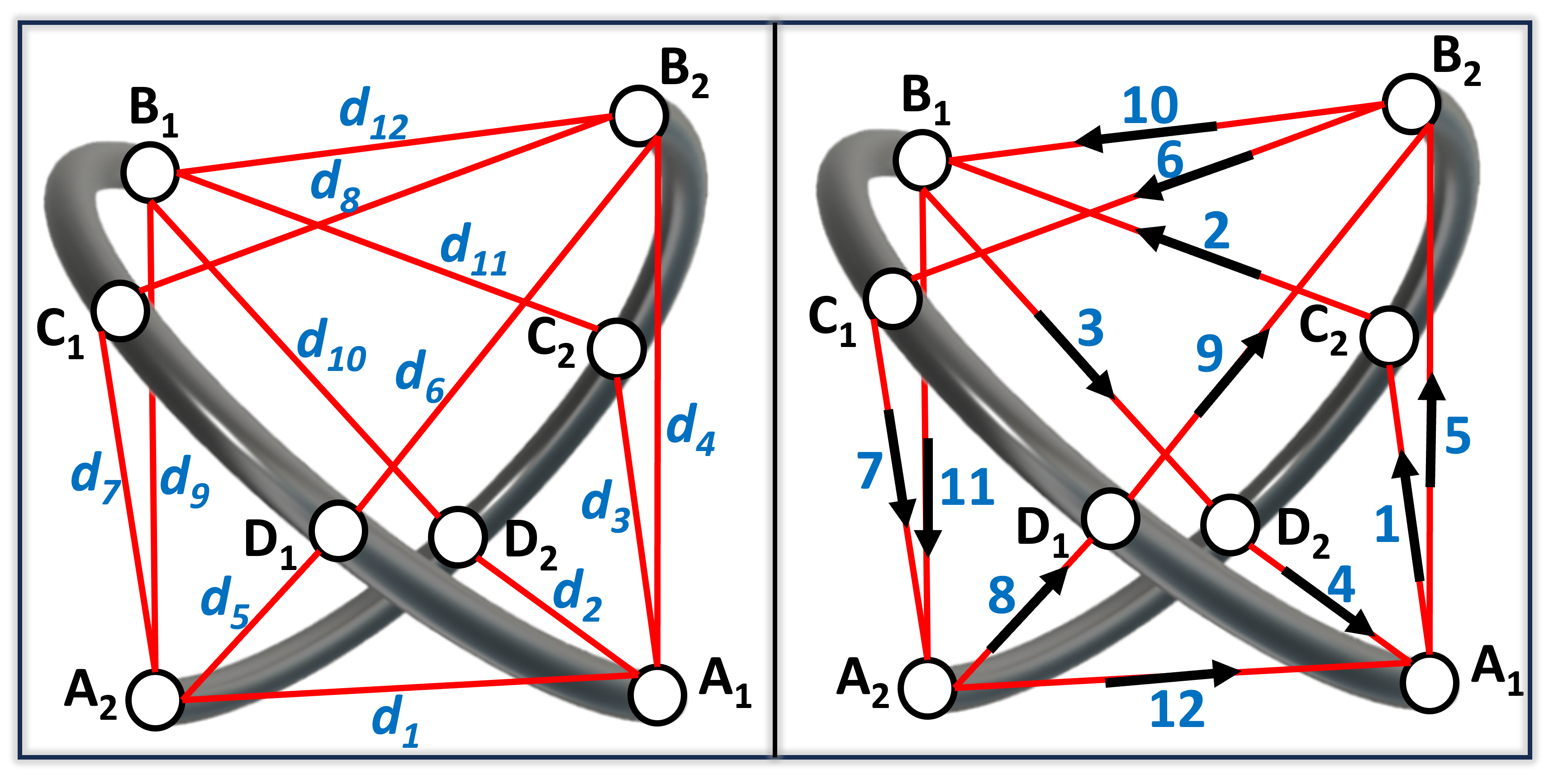}
    \subfloat(a){\hspace{.45\linewidth}}
    \subfloat(b){\hspace{.45\linewidth}}\vspace{-10pt}
    \caption{(a) The twelve cables joining different points on the two curved links are denoted by $\bm{d}_j, j\in[1,12]$. (b) The tendon routing path \EulerPath~ensures that each edge is only traversed once.}
    \label{fig:cab}
\end{figure}

Routing the cable through the two arcs such that the mechanism maintains structural integrity is challenging given the antagonistic nature of tensile cables and compressive rigid curved links. This is achieved by finding the Euler path of the graph where the connected edges and vertices correspond to the cables and connection points respectively. The reader may refer to \cite{woods_design_2023} for more details. One such Euler path, \EulerPath, is shown in \Fig \ref{fig:cab}(b). Here, the sequence simplifies the fabrication process by ensuring the cable traverses each edge only once. Once routed, the segments are tightened until the two curved links are held in tension with the optimal cable lengths identified through form-finding. 

\subsection{Mechatronics}

\begin{figure}[!ht]
    \centering
    \includegraphics[width = 0.8\columnwidth]{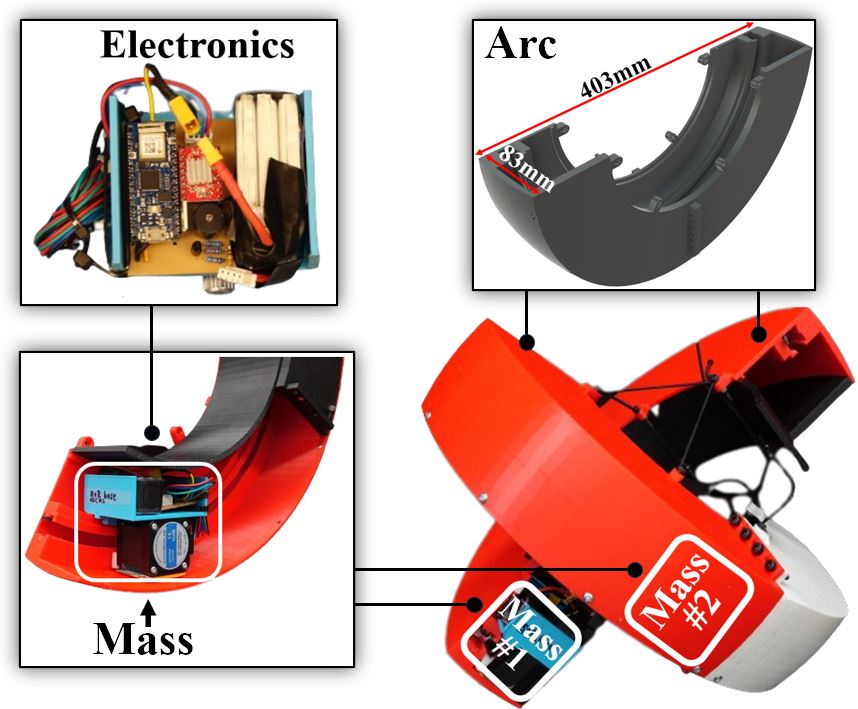}
    \caption{\Texplorer~comprises of two arcs with 83mm thickness and 403mm diameter. The internal shifting masses (mechatronics) that move along a GT2 belt internal to each arc.}
    \label{fig:fab}
\end{figure}

\Texplorer~is comprised of two 3D printed tough PLA curved links with 83mm thickness and 403mm diameter that are structurally held together with a cable using the discussed form-finding and routing approach. Motion is performed with a motor assembly traversing a GT2 timing belt running along the inside curvature of the arc. A side view of an open arc with the traveling motor assembly is shown in \Fig \ref{fig:fab}. The motor and electronic components are all positioned to one side of the timing belt with supporting delrin sliders on the top and side to maintain track alignment during movement. An Arduino Nano33 IoT, A4988 motor driver, 1,100mAh LiPo battery, and PCB are arranged compactly atop the traveling NEMA17 stepper motor. The assembly is designed with matching arc topology to reduce surface friction. Each semicircular arc weighs 431g and while each shifting mass weighs 427g, resulting in a nearly 1:1 weight ratio. The high ratio allows incremental movements of the masses to have a larger impact on the overall shifting center of mass. 

\section{RESULTS}
\subsection{Simulation}
\begin{figure}[htbp]
    \centering
    \includegraphics[width = 0.9\columnwidth]{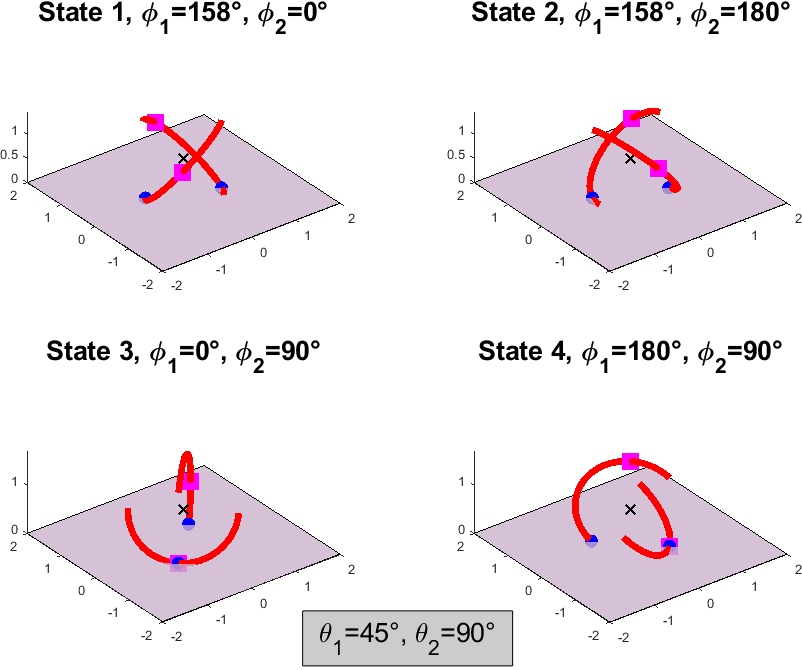}
    \caption{Four static equilibrium solutions for $(\theta_1 = 45 \degree, \theta_2 = 90 \degree)$. The blue dots are the points of contact $\bm{q}_i$, the pink squares are the positions of the internal masses $\bm{p}_i$, and the black `x' represents origin $O_b$ of the body coordinate system.}
    \label{fig:stat3}
\end{figure}

Simulations are performed in MATLAB\textregistered~to obtain the equilibrium position of the robot. Given the positions of the internal masses $(\theta_1,\theta_2)$, the equilibrium position can be interpreted as the two points of contacts represented through ground contact angles $(\phi_1,\phi_2)$. These four states of the robot are analytically found using \eqref{Eqn:State1}-\eqref{Eqn:State4}. As an example, \Fig \ref{fig:stat3} illustrates the four state solutions for internal mass positions $(\theta_1,\theta_2)=(45\degree,90\degree)$. Here, the black cross, pink squares and blue circles indicates the origin $O_b$ of the body coordinate system, internal masses and points of contact with the ground respectively. 

\begin{figure}
    \centering
    \includegraphics[width = \columnwidth]{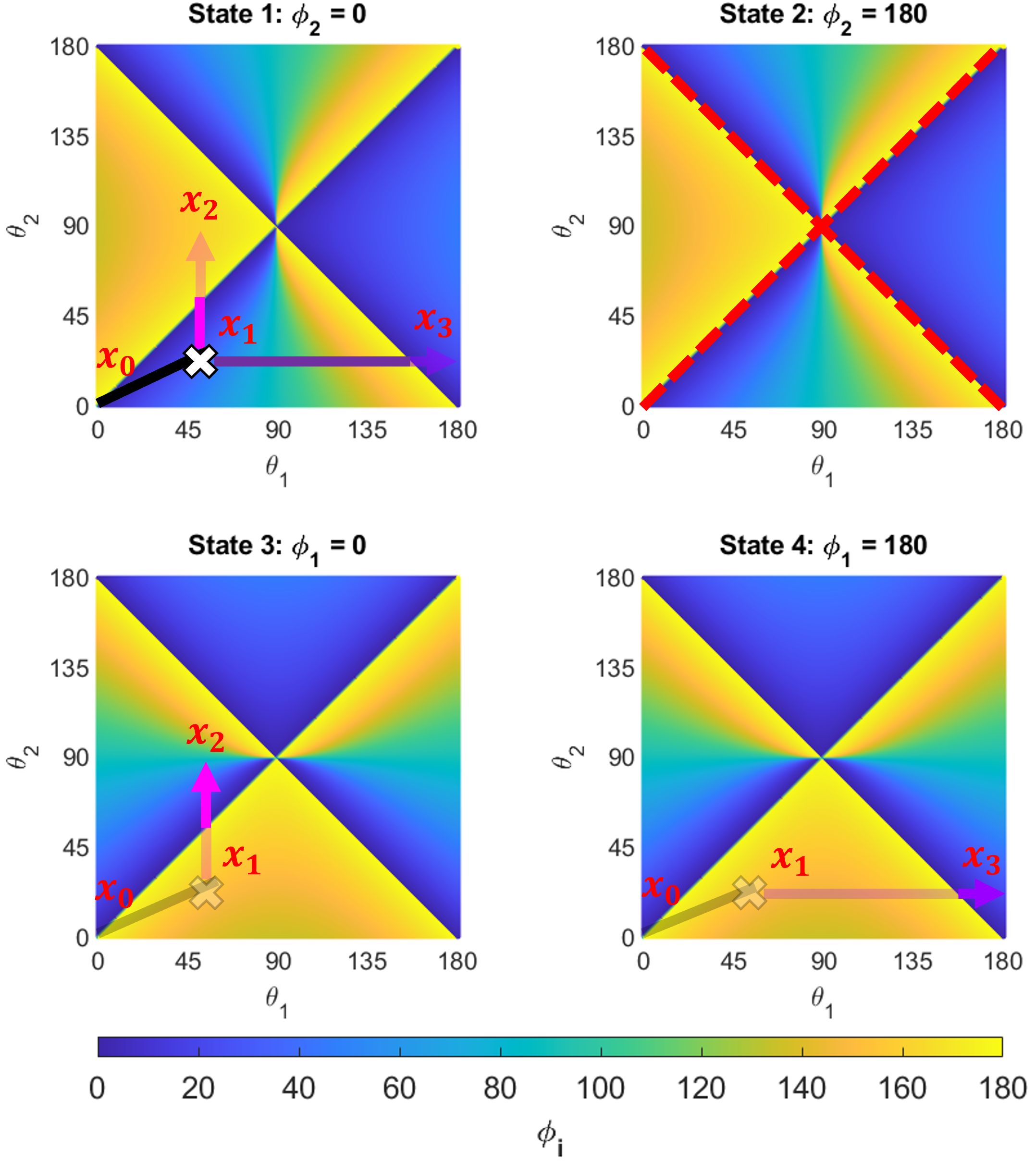}
    \caption{Simulation of static equilibrium positions of the four states. Two quasi-static control path sequences highlight the state transition boundaries (dotted red line): $x_0\rightarrow x_1 \rightarrow x_2$ for State $1 \rightarrow 3$ and $x_0\rightarrow x_1 \rightarrow x_3$ for State $1 \rightarrow 4$. }
    \label{fig:sims}
\end{figure}

The simulated ground contact angles for the four states as a function of the internal mass positions are shown in \Fig \ref{fig:sims}. As evident from the analytical solutions, the ground contact angles for States (1, 2) and (3, 4) solutions are identical. The plots demarcate the solutions in four distinct quadrants separated by the state transition boundaries highlighted by the dotted red lines in State 2 of \Fig \ref{fig:sims}. When considering this state, the bottom and top quadrants result in more movement of the robot, i.e., greater change in $\phi_2$ for change in $(\theta_1,\theta_2)$ as compared to the left and right quadrants. The state transition boundaries represent the internal mass positions where the robot can switch states. Consider two movement sequences visualized in  \Fig \ref{fig:sims} to highlight this phenomenon. The robot starts at internal mass position $x_0=(0,0)$ of State 1. Next, the internal masses are quasi-statically moved to position $x_1$ where $\phi_2$ is fixed at $0\degree$. Thereafter, the robot travels along a vertical path ($\theta_1$ constant, $\theta_2$ increasing) and approaches the state transition boundary where $\phi_1 \rightarrow 0\degree$. Instantaneously the robot enters the left quadrant of State 3 and $\phi_1$ begins to increase, and rolls to position $x_2$. The second movement again starts with the robot at position $x_0$ in State 1 rolling to $x_1$, and then begins to move towards the right state transition boundary as $\phi_1 \rightarrow 180\degree$. The robot transitions along the boundary and instantly enters the right quadrant of State 4, stopping at $x_3$.

\begin{figure*}
    \centering
    \includegraphics[width = 1.9\columnwidth]{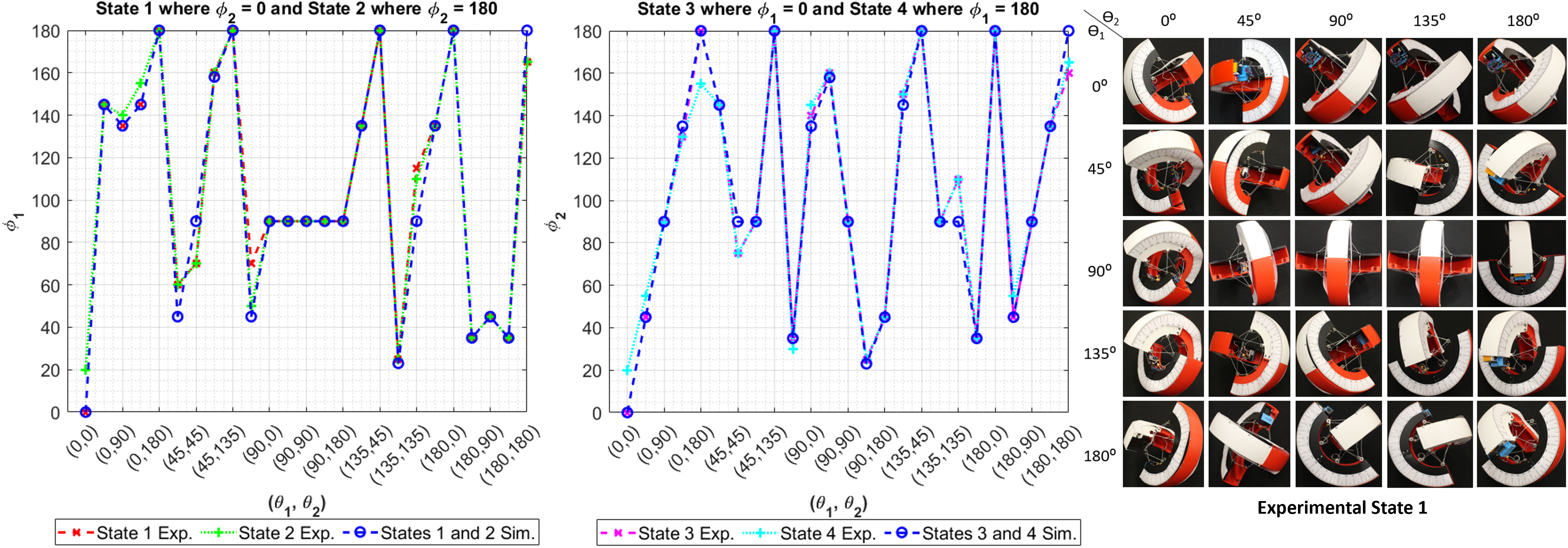}
    \begin{minipage}{0.66\textwidth}
    \begin{center}
        (a)
    \end{center}
    \end{minipage}
    \hfill
    \begin{minipage}{0.33\textwidth}
    \begin{center}
        (b)
    \end{center}
    \end{minipage}
    \caption{Experimental static solutions are validated through incrementally varying $(\theta_{1},\theta_2)$ by $45\degree$. (a) The experimental vs. simulated values for States 1- 4. (b) All overhead experimental results for State 1.}
    \label{fig:stability}
\end{figure*}

\subsection{Real-world experiments}

The prototype described in Sec. \ref{design} is used for experimental validation of the simulations. For combination of $(\theta_{1},\theta_2)$ in increments of $45 \degree$, the ground contact angles $(\phi_{1},\phi_2)$ corresponding to the robot orientation were observed. Protractor strips were attached along the curved links for observing the angles. All results were compared with simulation for each state in \Fig \ref{fig:stability}(a). For most of the inputs, the $\phi_i$ angles match exactly. 
The mean absolute error (MAE) over all the inputs is $4.36\degree$. The factors contributing to these differences between real-to-sim can be attributed to the assumptions and simplifications in the model of the system. More specifically, the curved links are modeled as 2D bars with a single ground contact point, however, the prototype has an arc thickness of $83mm$ with a curved arc endpoint allowing for a range of different ground contact points. Additionally, the center of mass of each shifting mass is modeled as a point mass while the mechatronics of the shifting mass are not centered due to uneven component layout. For example, the LiPo battery is attached to one side and weighs approximately a sixth of the total shifting mass. 
The experimental results for each input for State 1 are shown in \Fig \ref{fig:stability}(b). 

\section{CONCLUSIONS}

This research introduces a static modeling framework using geometric representation applicable to multi-point of contact systems. Here, the holonomic constraints prove the hybrid nature of the piecewise continuous rolling robot which exists in four states. Such multi-point contact system modeling has been explored for the first time in literature and the hybrid state model is verified by the real-world behavior of the robot. Each state physically corresponds to one of the end points of the two curved links about which the robot pivots while rolling along the other link. The modeling framework is generalizable and extendable to similar platforms of varying morphologies, i.e., robots with different shape, greater number of curved links, and varying curved link length, or their combination. The static analysis yields a four quadrant relationship between internal mass positions and ground contact angles. The quadrants are separated by state transition boundaries. Here, a quasi-static movement along these boundaries will result in change of states, i.e., physical change in the pivot points. The model is validated on a physical tetherless prototype with a MAE of $4.36\degree$, highlighting the accuracy of the model.


The future work involves dynamically modeling the \Texplorer~with consideration of dynamic movement of the internal mass. This will include a non-holonomic rolling constraint to account for pure rolling without slipping while constraining the solution space. 






\section*{ACKNOWLEDGMENT}
The authors would like to thank Chase Fortin for his help in fabricating several iterations of the prototype.


\appendices

\bibliographystyle{IEEEtran}
\bibliography{IEEEabrv,texplorZ,texplor}

\begin{thebibliography}{10}
\providecommand{\url}[1]{#1}
\csname url@samestyle\endcsname
\providecommand{\newblock}{\relax}
\providecommand{\bibinfo}[2]{#2}
\providecommand{\BIBentrySTDinterwordspacing}{\spaceskip=0pt\relax}
\providecommand{\BIBentryALTinterwordstretchfactor}{4}
\providecommand{\BIBentryALTinterwordspacing}{\spaceskip=\fontdimen2\font plus
\BIBentryALTinterwordstretchfactor\fontdimen3\font minus
  \fontdimen4\font\relax}
\providecommand{\BIBforeignlanguage}[2]{{%
\expandafter\ifx\csname l@#1\endcsname\relax
\typeout{** WARNING: IEEEtran.bst: No hyphenation pattern has been}%
\typeout{** loaded for the language `#1'. Using the pattern for}%
\typeout{** the default language instead.}%
\else
\language=\csname l@#1\endcsname
\fi
#2}}
\providecommand{\BIBdecl}{\relax}
\BIBdecl

\bibitem{gim_ringbot_2024}
\BIBentryALTinterwordspacing
K.~G. Gim and J.~Kim, ``Ringbot: {Monocycle} {Robot} {With} {Legs},''
  \emph{IEEE Transactions on Robotics}, vol.~40, pp. 1890--1905, 2024,
  conference Name: IEEE Transactions on Robotics. [Online]. Available:
  \url{https://ieeexplore.ieee.org/document/10423226}
\BIBentrySTDinterwordspacing

\bibitem{saranli2001rhex}
U.~Saranli, M.~Buehler, and D.~E. Koditschek, ``Rhex: A simple and highly
  mobile hexapod robot,'' \emph{The International Journal of Robotics
  Research}, vol.~20, no.~7, pp. 616--631, 2001.

\bibitem{quinn2003parallel}
R.~D. Quinn, G.~M. Nelson, R.~J. Bachmann, D.~A. Kingsley, J.~T. Offi, T.~J.
  Allen, and R.~E. Ritzmann, ``Parallel complementary strategies for
  implementing biological principles into mobile robots,'' \emph{The
  International Journal of Robotics Research}, vol.~22, no. 3-4, pp. 169--186,
  2003.

\bibitem{chase_review_2012}
\BIBentryALTinterwordspacing
R.~Chase and A.~Pandya, ``\BIBforeignlanguage{en}{A {Review} of {Active}
  {Mechanical} {Driving} {Principles} of {Spherical} {Robots}},''
  \emph{\BIBforeignlanguage{en}{Robotics}}, vol.~1, no.~1, pp. 3--23, Dec.
  2012, number: 1 Publisher: Multidisciplinary Digital Publishing Institute.
  [Online]. Available: \url{https://www.mdpi.com/2218-6581/1/1/3}
\BIBentrySTDinterwordspacing

\bibitem{morinaga_motion_2014}
A.~Morinaga, M.~Svinin, and M.~Yamamoto, ``A {Motion} {Planning} {Strategy} for
  a {Spherical} {Rolling} {Robot} {Driven} by {Two} {Internal} {Rotors},''
  \emph{IEEE Transactions on Robotics}, vol.~30, no.~4, pp. 993--1002, Aug.
  2014.

\bibitem{ohsawa_geometric_2020}
\BIBentryALTinterwordspacing
T.~Ohsawa, ``\BIBforeignlanguage{en}{Geometric {Kinematic} {Control} of a
  {Spherical} {Rolling} {Robot}},'' \emph{\BIBforeignlanguage{en}{Journal of
  Nonlinear Science}}, vol.~30, no.~1, pp. 67--91, Feb. 2020. [Online].
  Available: \url{http://link.springer.com/10.1007/s00332-019-09568-x}
\BIBentrySTDinterwordspacing

\bibitem{snelson1996snelson}
K.~Snelson, ``Snelson on the tensegrity invention,'' \emph{International
  Journal of Space Structures}, vol.~11, no. 1-2, pp. 43--48, 1996.

\bibitem{skelton2009tensegrity}
R.~E. Skelton and M.~C. De~Oliveira, \emph{Tensegrity systems}.\hskip 1em plus
  0.5em minus 0.4em\relax Springer, 2009, vol.~1.

\bibitem{paul_design_2006}
C.~Paul, F.~Valero-Cuevas, and H.~Lipson, ``Design and control of tensegrity
  robots for locomotion,'' \emph{IEEE Transactions on Robotics}, vol.~22,
  no.~5, pp. 944--957, Oct. 2006.

\bibitem{sabelhaus_etal_icra_2015}
\BIBentryALTinterwordspacing
A.~P. Sabelhaus, J.~Bruce, K.~Caluwaerts, P.~Manovi, R.~F. Firoozi, S.~Dobi,
  A.~M. Agogino, and V.~SunSpiral, ``System design and locomotion of
  {SUPERball}, an untethered tensegrity robot,'' in \emph{2015 {IEEE}
  {International} {Conference} on {Robotics} and {Automation} ({ICRA})}.\hskip
  1em plus 0.5em minus 0.4em\relax Seattle, WA, USA: IEEE, May 2015, pp.
  2867--2873. [Online]. Available:
  \url{http://ieeexplore.ieee.org/document/7139590/}
\BIBentrySTDinterwordspacing

\bibitem{caluwaerts_etal_royal_society_interface_2014}
\BIBentryALTinterwordspacing
K.~Caluwaerts, J.~Despraz, A.~Işçen, A.~P. Sabelhaus, J.~Bruce, B.~Schrauwen,
  and V.~SunSpiral, ``\BIBforeignlanguage{en}{Design and control of compliant
  tensegrity robots through simulation and hardware validation},''
  \emph{\BIBforeignlanguage{en}{Journal of The Royal Society Interface}},
  vol.~11, no.~98, p. 20140520, Sep. 2014. [Online]. Available:
  \url{https://royalsocietypublishing.org/doi/10.1098/rsif.2014.0520}
\BIBentrySTDinterwordspacing

\bibitem{chen_soft_2017}
\BIBentryALTinterwordspacing
L.-H. Chen, K.~Kim, E.~Tang, K.~Li, R.~House, E.~L. Zhu, K.~Fountain, A.~M.
  Agogino, A.~Agogino, V.~Sunspiral, and E.~Jung, ``Soft {Spherical}
  {Tensegrity} {Robot} {Design} {Using} {Rod}-{Centered} {Actuation} and
  {Control},'' \emph{Journal of Mechanisms and Robotics}, vol.~9, no. 025001,
  Mar. 2017. [Online]. Available: \url{https://doi.org/10.1115/1.4036014}
\BIBentrySTDinterwordspacing

\bibitem{rieffel2018adaptive}
J.~Rieffel and J.-B. Mouret, ``Adaptive and resilient soft tensegrity robots,''
  \emph{Soft robotics}, vol.~5, no.~3, pp. 318--329, 2018.

\bibitem{rhodes_compact_2019}
\BIBentryALTinterwordspacing
T.~Rhodes, C.~Gotberg, and V.~Vikas, ``Compact {Shape} {Morphing} {Tensegrity}
  {Robots} {Capable} of {Locomotion},'' \emph{Frontiers in Robotics and AI},
  vol.~6, 2019. [Online]. Available:
  \url{https://www.frontiersin.org/articles/10.3389/frobt.2019.00111}
\BIBentrySTDinterwordspacing

\bibitem{bohm_spherical_2016}
V.~Böhm, T.~Kaufhold, F.~Schale, and K.~Zimmermann, ``Spherical mobile robot
  based on a tensegrity structure with curved compressed members,'' in
  \emph{2016 {IEEE} {International} {Conference} on {Advanced} {Intelligent}
  {Mechatronics} ({AIM})}, Jul. 2016, pp. 1509--1514.

\bibitem{bohm_dynamic_2017}
\BIBentryALTinterwordspacing
V.~Böhm, T.~Kaufhold, I.~Zeidis, and K.~Zimmermann,
  ``\BIBforeignlanguage{en}{Dynamic analysis of a spherical mobile robot based
  on a tensegrity structure with two curved compressed members},''
  \emph{\BIBforeignlanguage{en}{Archive of Applied Mechanics}}, vol.~87, no.~5,
  pp. 853--864, May 2017. [Online]. Available:
  \url{https://doi.org/10.1007/s00419-016-1183-z}
\BIBentrySTDinterwordspacing

\bibitem{kaufhold_indoor_2017}
T.~Kaufhold, F.~Schale, V.~Böhm, and K.~Zimmermann, ``Indoor locomotion
  experiments of a spherical mobile robot based on a tensegrity structure with
  curved compressed members,'' in \emph{2017 {IEEE} {International}
  {Conference} on {Advanced} {Intelligent} {Mechatronics} ({AIM})}, Jul. 2017,
  pp. 523--528, iSSN: 2159-6255.

\bibitem{schorr_kinematic_2021}
\BIBentryALTinterwordspacing
P.~Schorr, E.~R.~C. Li, T.~Kaufhold, J.~A.~R. Hernández, L.~Zentner,
  K.~Zimmermann, and V.~Böhm, ``\BIBforeignlanguage{en}{Kinematic analysis of
  a rolling tensegrity structure with spatially curved members},''
  \emph{\BIBforeignlanguage{en}{Meccanica}}, vol.~56, no.~4, pp. 953--961, Apr.
  2021. [Online]. Available: \url{https://doi.org/10.1007/s11012-020-01199-x}
\BIBentrySTDinterwordspacing

\bibitem{antali_slippingrolling_2022}
\BIBentryALTinterwordspacing
M.~Antali and G.~Stepan, ``\BIBforeignlanguage{en}{Slipping–rolling
  transitions of a body with two contact points},''
  \emph{\BIBforeignlanguage{en}{Nonlinear Dynamics}}, vol. 107, no.~2, pp.
  1511--1528, Jan. 2022. [Online]. Available:
  \url{https://doi.org/10.1007/s11071-021-06538-5}
\BIBentrySTDinterwordspacing

\bibitem{murray_mathematical_2017}
\BIBentryALTinterwordspacing
R.~M. Murray, Z.~Li, and S.~S. Sastry, \emph{\BIBforeignlanguage{en}{A
  {Mathematical} {Introduction} to {Robotic} {Manipulation}}}, 1st~ed.\hskip
  1em plus 0.5em minus 0.4em\relax CRC Press, Dec. 2017. [Online]. Available:
  \url{https://www.taylorfrancis.com/books/9781351469791}
\BIBentrySTDinterwordspacing

\bibitem{tibert_review_2003}
\BIBentryALTinterwordspacing
A.~Tibert and S.~Pellegrino, ``\BIBforeignlanguage{en}{Review of
  {Form}-{Finding} {Methods} for {Tensegrity} {Structures}},''
  \emph{\BIBforeignlanguage{en}{International Journal of Space Structures}},
  vol.~18, no.~4, pp. 209--223, Dec. 2003, publisher: SAGE Publications Ltd
  STM. [Online]. Available: \url{https://doi.org/10.1260/026635103322987940}
\BIBentrySTDinterwordspacing

\bibitem{connelly_second-order_1996}
\BIBentryALTinterwordspacing
R.~Connelly and W.~Whiteley, ``Second-{Order} {Rigidity} and {Prestress}
  {Stability} for {TensegrityFrameworks},'' \emph{SIAM Journal on Discrete
  Mathematics}, vol.~9, no.~3, pp. 453--491, Aug. 1996. [Online]. Available:
  \url{https://doi.org/10.1137/S0895480192229236}
\BIBentrySTDinterwordspacing

\bibitem{woods_design_2023}
\BIBentryALTinterwordspacing
C.~Woods and V.~Vikas, ``Design and {Modeling} {Framework} for {DexTeR}:
  {Dexterous} {Continuum} {Tensegrity} {Manipulator},'' \emph{Journal of
  Mechanisms and Robotics}, vol.~15, no. 031006, Mar. 2023. [Online].
  Available: \url{https://doi.org/10.1115/1.4056959}
\BIBentrySTDinterwordspacing

\end{thebibliography}
\addtolength{\textheight}{-12cm}

\end{document}